\documentclass[11pt,a4paper]{article}
\usepackage{amsfonts}
\usepackage{amsmath}
\usepackage{bm}
\usepackage[hyperref]{eacl2021}
\usepackage{multirow}
\usepackage{times}
\usepackage{latexsym}
\usepackage{float}

\usepackage{tikz}
\usepackage[nottoc]{tocbibind}
\usetikzlibrary{shapes.multipart}
\usetikzlibrary{positioning}
\usetikzlibrary{arrows.meta}
\usetikzlibrary{backgrounds}

\newcommand{\pr}{\mathbb{P}}

\newcommand{\roughly}{\raise.17ex\hbox{$\scriptstyle\sim$}}
\newcommand{\pd}[2]{\frac{\partial #1}{\partial #2}}

\usepackage{microtype}

\aclfinalcopy 

\usepackage{graphicx}
\graphicspath{ {images/} }
\usepackage{caption}
\usepackage{subcaption}

\usepackage{xcolor}
\usepackage{pbox}

\title{PolyLM: Learning about Polysemy through Language Modeling}

\author{Alan Ansell$^{1,2}$\ , Felipe Bravo-Marquez$^3$, Bernhard Pfahringer$^2$\\
        $^1$Language Technology Lab, University of Cambridge\\
        $^2$Department of Computer Science, University of Waikato\\
        $^3$Department of Computer Science, University of Chile \& IMFD\\
}

\date{}

\begin{document}
\maketitle

\begin{abstract}
To avoid the ``meaning conflation deficiency'' of word embeddings, a number of models have aimed to embed individual word \textit{senses}. These methods at one time performed well on tasks such as word sense induction (WSI), but they have since been overtaken by task-specific techniques which exploit contextualized embeddings. However, sense embeddings and contextualization need not be mutually exclusive. We introduce PolyLM, a method which formulates the task of learning sense embeddings as a language modeling problem, allowing contextualization techniques to be applied. PolyLM is based on two underlying assumptions about word senses: firstly, that the probability of a word occurring in a given context is equal to the sum of the probabilities of its individual senses occurring; and secondly, that for a given occurrence of a word, one of its senses tends to be much more plausible in the context than the others. We evaluate PolyLM on WSI, showing that it performs considerably better than previous sense embedding techniques, and matches the current state-of-the-art specialized WSI method despite having six times fewer parameters. Code and pre-trained models are available at \url{https://github.com/AlanAnsell/PolyLM}.
\end{abstract}

\section{Introduction}

Much work in NLP has been dedicated to vector representations of words, but it has been recognized since as early as \cite{schutze-1998-automatic} that such representations fail to capture the polysemous nature of many words, conflating their multiple senses into a single point in semantic space. There have been several attempts at embedding individual word senses to avoid this issue, termed the ``meaning conflation deficiency'' by \citet{camacho-collados-2018-from} in their survey on the area.

We propose PolyLM, an unsupervised sense embedding model which is effective and easy to apply to downstream tasks. PolyLM can be thought of as both a (masked) language model and a \textit{sense} model, as it calculates a probability distribution both over words and word senses at masked positions. The formulation is derived from two observations about word senses: firstly, that the probability of a word occurring in a given context is equal to the sum of the probabilities of its individual senses occurring; and secondly, that for a given occurrence of a word, one of its senses tends to be much more plausible in the context than the others.

\begin{figure*}[htb]
	\captionsetup[subfigure]{justification=centering}
	\centering
	\begin{subfigure}{0.45\linewidth}
		\centering
		\fbox{\includegraphics[width=0.98\linewidth]{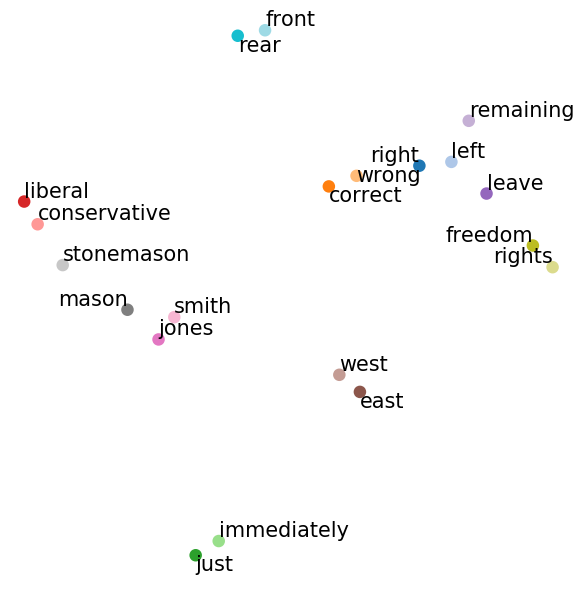}}
		\caption{Word embeddings}
		\label{fig:word embeddings}
	\end{subfigure}
	\begin{subfigure}{0.45\linewidth}
		\centering
		\fbox{\includegraphics[width=0.98\linewidth]{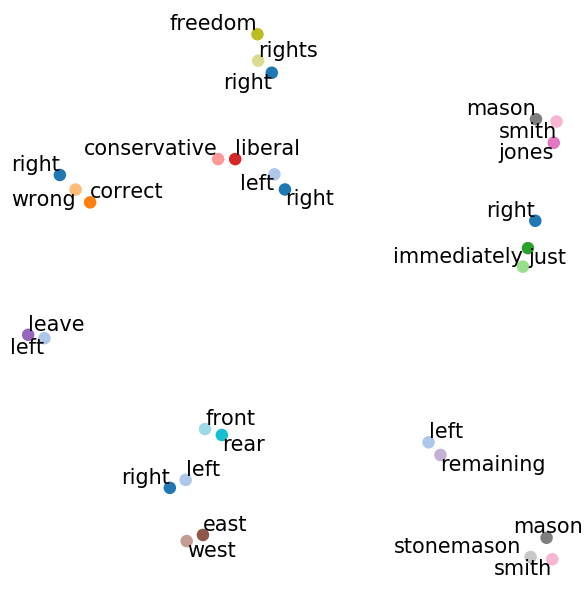}}
		\caption{Sense embeddings}
		\label{fig:sense embeddings}
	\end{subfigure}
	\caption{An illustration of the meaning conflation deficiency, showing selected word and sense embeddings learned by PolyLM visualized using t-SNE \cite{maaten-hinton-2008-visualizing} and adjustText \cite{adjust-text}. Sense embeddings were learned by training PolyLM$_\text{SMALL}$ with the standard 8 senses per word; word embeddings were learned by training PolyLM$_\text{SMALL}$, but with a single sense per word. Note that both models were trained on unlemmatized data, unlike those used in the WSI experiments. The occurrence of closely related polysemous words nearby in the word embedding space (i.e. \textit{left} and \textit{right}) causes unrelated words to be closer together (e.g. \textit{left} and \textit{wrong}) and related words to be further apart (e.g. \textit{right} and \textit{east}) than they otherwise would be. The use of sense embeddings avoids such distortion. PolyLM is capable of detecting comparatively rare word senses, such as the political senses of \textit{left} and \textit{right}, and the use of \textit{smith} and \textit{mason} to refer to tradespeople.}
	\label{fig:embeddings plot}
\end{figure*}

There are several reasons for the interest in sense representations. The first is the downsides associated with the meaning conflation deficiency. Word embedding models can have difficulty distinguishing which sense of an ambiguous word applies in a given context \cite{yaghoobzadeh-schutze-2016-intrinsic}. Additionally, homonymy and polysemy cause distortion in word embeddings: for instance, we would find the unrelated words \textit{left} and \textit{wrong} unreasonably close in the vector space due to their similarity to two different senses of the word \textit{right}, an effect noted by \citet{neelakantan-etal-2014-efficient} and illustrated in Figure \ref{fig:embeddings plot}. Intuitively we would expect that sense embedding models could gain superior semantic understanding by avoiding these problems.

\setlength{\fboxsep}{0pt}

In addition to well-established applications for sense representations such as word sense disambiguation (WSD) and induction (WSI), another interesting use case is the automatic construction of lexical resources \cite{neale-2018-survey}. While there are existing human-curated word sense inventories for English such as such as WordNet \cite{miller-1995-wordnet}, these are expensive to create and are unavailable for most languages. \citet{panchenko-2016-best} showed that sense embeddings learned using the model of \citet{bartunov-etal-2016-breaking} could be linked with word senses contained in BabelNet \cite{navigli-ponzetto-2012-babelnet} with a reasonable degree of precision, although the mapping struggled with recall. PolyLM represents a significant advance over \citeauthor{bartunov-etal-2016-breaking}'s in terms of WSI performance, so it seems reasonable to imagine that this approach to lexical resource construction might now be more feasible.

The emergence of contextualized models such as ELMo \cite{peters-etal-2018-deep} and BERT \cite{devlin-etal-2019-bert} has had a tremendous impact on the area of semantic representation. Rather than representing words using a single embedding, or even a set of sense embeddings, these models allow words to be represented using an infinite set of possible embeddings depending on the context. This approach has been very effective across NLP and many state-of-the-art systems incorporate contextualized models, including systems for WSD and WSI. The success of contextualized models raises the question of whether there is still value in learning discrete sense representations.

However, contextualized models still rely on word embeddings, and are therefore subject to the meaning conflation deficiency. Furthermore it could be argued that it is inefficient to have the same representation size for all words regardless of how diverse their range of senses is. Another drawback is that before they can be applied to word sense-related tasks, an adaptation step such as clustering to induce discrete senses or fine-tuning is generally required, which is often expensive in terms of both research and compute time.

The contributions of this paper can be summarized as follows:
\begin{itemize}
    \item We propose PolyLM, an end-to-end, unsupervised neural sense embedding model derived from two simple assumptions about word senses. We demonstrate that PolyLM learns senses which correspond well to human notions by showing that it performs well at WSI.
    \item PolyLM is flexible in that it can use any ``contextualizer'' (a useful term coined by \citet{liu-etal-2019-linguistic}), so it will remain relevant as contextualization techniques improve.
    \item We reduce the effect of the meaning conflation deficiency by disambiguating word senses at the input with a neural ``disambiguation layer.'' We show that good performance on WSI can be achieved using the output of this layer alone, suggesting that it could be a useful component in many neural networks for language understanding.
\end{itemize}

\section{Related Work}
One of the first works in unsupervised learning of sense representations was by \citet{schutze-1998-automatic}, who proposed a two-step process, where vector representations are first derived for each context containing an ambiguous word, and these are then clustered into a pre-defined number of groups. \citet{huang-etal-2012-improving} added a third step, where after sense-labeling each word according to its context cluster, sense representations are learned through neural language modeling.

A number of later approaches employed a joint training approach, where sense labeling and sense representation learning happen in parallel. \citet{neelakantan-etal-2014-efficient}, \citet{li-jurafsky-2015-multi} and \citet{bartunov-etal-2016-breaking} each proposed multi-sense variants of the Skip-Gram model \cite{mikolov-etal-2013-efficient}. Various approaches were tried for determining the number of senses per word: for instance, \citeauthor{li-jurafsky-2015-multi} and \citeauthor{bartunov-etal-2016-breaking} used Chinese Restaurant Processes and Dirichlet Processes respectively to automatically learn an appropriate number of senses for each word.

Many joint training approaches have the disadvantage that they create ambiguity in the context representation by representing context words with word embeddings in order to avoid considering the exponential number of possible sense labelings for the context. \citet{qiu-etal-2016-context} and \citet{lee-chen-2017-muse} propose purely sense-based approaches which can sense-label the input efficiently.

\citet{arora-etal-2018-linear} took a novel approach to the problem of learning word senses, demonstrating that the embedding learned by traditional techniques for an ambiguous word tends to be very close to a linear combination of the hypothetical vectors corresponding to its individual senses. They proposed a method for recovering the underlying sense vectors and coefficients, and evaluated their system on WSI.

Since the emergence of contextualized models, there have been a number of other systems which have exploited their powerful semantic representations for specific tasks such as word sense disambiguation \cite{huang-etal-2019-glossbert, vial-etal-2019-sense} and induction \cite{amrami-goldberg-2018-word,amrami-goldberg-2019-towards}, however none of these methods creates explicit sense embeddings.

\section{PolyLM} \label{sec:model}

\subsection{Overview}
Consider a typical neural language model. Each word $w$ in a vocabulary $V$ is assigned a single embedding, resulting in an embedding matrix $M \in \mathbb{R}^{|V| \times d}$, where $d$ is the embedding dimensionality. The probability of $w$ occurring in a context $c$ is estimated as
\begin{align} \label{eq:standard lm objective}
    \pr(w\ |\ c) = \big[\text{softmax}\big(M \bm{y}(c) + \bm{a}\big)\big]_w,
\end{align}
where $\bm{y}(c) \in \mathbb{R}^d$ is a vector representation of $c$ and $\bm{a} \in \mathbb{R}^{|V|}$ is a trainable bias vector. In BERT \cite{devlin-etal-2019-bert} for instance, $\bm{y}(c)$ corresponds to the final output of multiple Transformer encoder layers \cite{vaswani-etal-2017-attention}.

Now suppose that for each $w \in V$, there is a corresponding set $S_w$ of \textit{sememes}, or senses which $w$ can have. For instance, intuitively we might have $S_\text{rock} = \{\text{rock:stone, rock:musical genre, rock:shake}\}$. We assume that the $S_w$ are disjoint, i.e. $S_w \cap S_{w'} = \emptyset$ whenever $w \neq w'$, and we define the full sense inventory $S = \bigcup_{w \in V} S_w$.

Context induces specific senses for the words it contains. Thus a passage of text can be thought of as a sequence of sememes as well as a sequence of words. The first observation underlying PolyLM is that the probability of a word $w$ occurring in a context $c$ is equal to the sum of the probabilities of $w$'s component sememes occurring in the context, i.e.
\begin{align} \label{eq:senses to words}
    \pr(w\ |\ c) = \sum_{s \in S_w} \pr(s\ |\ c).
\end{align}

We wish to learn representations for individual senses, and so we assign an embedding to each sememe in our sense inventory, resulting in a matrix $E$ with dimension $|S| \times d$ and bias vector $\bm{b}$ of dimension $|S|$. Note that this assumes that we know the number of senses of each word \textit{a priori}, an assumption whose consequences we discuss later. Following Eq. \ref{eq:standard lm objective}, we define the vector $\bm{p}(c) \in \mathbb{R}^{|S|}$ of sememe probabilities in a context $c$ as
\begin{align} \label{eq:sense probability vector}
    \bm{p}(c) = \text{softmax}\big(E \bm{x}(c) + \bm{b}\big).
\end{align}
Considering Eq. \ref{eq:senses to words}, we have
\begin{align} \label{eq:objective overview}
    \pr(w\ |\ c) = \sum_{s \in S_w} p(c)_s,
\end{align}
allowing us to formulate the problem of learning sense representations with a language modeling objective.

PolyLM is constructed from three components: the input layer, which represents the input tokens as aggregates of their sense embeddings, the disambiguation layer, which attempts to determine the contextually appropriate sense embeddings for the input, and the prediction layer, which implements the language modeling objective.

We adopt the masked language modeling (MLM) task used for training BERT. When training, we select a subset $T \subset \{1, 2, ..., n\}$ of the tokens in the input sequence as targets for prediction, and produce a masked version $c' = w'_1, w'_2,..., w'_n$ of the original sequence $c = w_1, w_2, ..., w_n$ as follows: 15\% of tokens are chosen at random as targets, of which 80\% are replaced with a special [MASK] token, 10\% are replaced with a random token, and 10\% are left unchanged.

\newcommand{\horsp}{22mm}
\newcommand{\minw}{15mm}
\newcommand{\vertsp}{15mm}
\newcommand{\minh}{7mm}
\setlength{\fboxsep}{7pt}
\begin{figure*}[p]
    \centering
    \resizebox{0.9\linewidth}{!}{\begin{tikzpicture}[
        framed,
        inner frame xsep=3mm,
        inner frame ysep=5mm,
        background rectangle/.style={
            ultra thick,
            rounded corners=3mm, 
            draw=black,
            fill={rgb,255:red,244; green,241; blue,222}
        },
        single/.style={
            rectangle,
            draw,
            very thick,
            rounded corners=2mm,
            minimum width=\minw,
            minimum height=\minh,
            node distance=5mm
        },
        token/.style={
            rectangle,
            draw,
            very thick,
            rounded corners=2mm,
            minimum width=\minw,
            minimum height=\minh,
            node distance=5mm,
            fill={rgb,255:red,129; green,178; blue,154},
            font=\ttfamily
        },
        vec/.style={
            rectangle,
            draw,
            very thick,
            rounded corners=2mm,
            minimum width=\minw,
            minimum height=\minh,
            node distance=5mm,
            fill={rgb,255:red,105; green,183; blue,201},
        },
        loss/.style={
            rectangle,
            draw,
            very thick,
            rounded corners=2mm,
            minimum width=\minw,
            minimum height=\minh,
            node distance=5mm,
            fill={rgb,255:red,224; green,122; blue,95},
        },
        prob/.style={
            rectangle,
            draw,
            very thick,
            rounded corners=2mm,
            minimum width=\minw,
            minimum height=\minh,
            node distance=5mm,
            fill={rgb,255:red,242; green,204; blue,143},
            font=\sffamily,
        },
        transformer/.style={
            rectangle,
            draw,
            rounded corners=3mm,
            minimum width=5*\horsp,
            minimum height=15mm,
            node distance=5mm,
            very thick,
            fill={rgb,255:red,177; green,150; blue,125},
            font=\sffamily
        },
        arrow/.style={
            ->,
            very thick,
        },
        comment/.style={
            font=\sffamily,
        }
        ]
        
        \foreach \i [count=\step from 1] in {i,like,apple,pie,.} {
            \node [token] (os\step) at (\step*\horsp, 0) {\i};
        }
        \node[comment] [left=1mm of os1] {Unmasked sequence $c$};
        
        \foreach \i [count=\step from 1] in {i,like,[MASK],pie,.} {
            \node [token] (ms\step) [below=of os\step] {\i};
            \draw[arrow] (os\step.south) -> (ms\step.north);
        }
        \node[comment] [left=1mm of ms1] {Masked sequence $c'$};
        
        \foreach \step in {1,...,5} {
            \node [vec] (ie\step) [below=of ms\step] {$\bm{x}(w'_{\step})$};
            \draw[arrow] (ms\step.south) -> (ie\step.north);
        }
        \node[comment] [left=1mm of ie1] {Input embeddings for $c'$};
        
        \node[transformer] (dlayer1) [below=of ie3] {Disambiguation Transformer Encoder, $C^D$};
        
        \foreach \step in {1,...,5} {
            \draw[arrow] (ie\step.south) -> (dlayer1.north -| ie\step.south);
            \node [vec] (do\step) [below=of dlayer1.south -| ms\step] {$\bm{y}_{\step}^D(c')$};
            \draw[arrow] (dlayer1.south -| do\step) -> (do\step.north);
            \node [vec] (dr\step) [below=of do\step] {$\bm{x}_{\step}^P(c')$};
            \draw[arrow] (do\step.south) -> (dr\step.north);
        }
        \node[comment] [left=1mm of do1] {
            \begin{tabular}{c}
                Disambiguation\\
                output for $c'$\\
            \end{tabular}
        };
        \node[comment] [left=1mm of dr1] {
            \begin{tabular}{c}
                Disambiguated\\
                input representations\\
            \end{tabular}
        };
        
        \node[transformer] (player) [below=of dr3] {Prediction Transformer Encoder, $C^P$};
        
        \foreach \step in {1,...,5} {
            \draw[arrow] (dr\step.south) -> (player.north -| dr\step.south);
            \node [vec] (pr\step) [below=of player.south -| ms\step] {$\bm{y}_{\step}^P(c')$};
            \draw[arrow] (player.south -| pr\step) -> (pr\step.north);
        }
        \node[comment] [left=1mm of pr1] {Output representations};
        
        \node[prob] (p3) [below=of pr3] {
            \begin{tabular}{rl}
                \multicolumn{2}{c}{$\bm{p}_3(c')$}\\
                \multicolumn{2}{c}{.}\\
                \multicolumn{2}{c}{.}\\
                \texttt{apple}$_1$: & 0.00006\\
                \texttt{apple}$_2$: & 0.05164\\
                \texttt{apple}$_3$: & 0.00012\\
                \multicolumn{2}{c}{.}\\
                \multicolumn{2}{c}{.}\\
            \end{tabular}
        };
        \draw[arrow] (pr3.south) -> (p3.north);
        \node[comment] [left=1mm of p3 -| pr1.west] {
            \begin{tabular}{c}
                Sense probabilities\\
                summed to give word\\
                probabilities for\\
                masked words, used\\
                to calculate language\\
                modeling loss $J^{LM}$\\
            \end{tabular}
        };
        
        \node[prob] (totalp) [right=of p3.north -| p3.east, anchor=north west] {
            \begin{tabular}{l}
                $\pr(\text{\texttt{apple}}) =$\\
                \ 0.00006 +\\
                \ \ 0.05164 +\\
                \ \ 0.00012\\
                \ = 0.0518\\
            \end{tabular}
        };
        \draw[arrow] (p3.east |- totalp.west) -> (totalp.west);
        \node[loss] (jlm) [below=of totalp] {$J^{LM}$};
        \draw[arrow] (totalp.south) -> (jlm.north);

        \node[prob] (q3p) [below=of p3.south -| pr1.west, anchor=north west] {
            \begin{tabular}{rl}
                \multicolumn{2}{c}{$\bm{q}_3^P(c', c)$}\\
                \texttt{apple$_1$}: & 0.0011\\
                \texttt{apple$_2$}: & 0.9966\\
                \texttt{apple$_3$}: & 0.0022\\
            \end{tabular}
        };
        \node[shape=coordinate] (int1) [below=2.5mm of pr3.south] {};
        \draw (int1) [rounded corners] -- (int1 -| q3p) [arrow] -- (q3p.north);
        
        \foreach \step in {1,...,5} {
            \node [vec] (ddr\step) [below=of q3p.south-|ms\step] {$\bm{y}_{\step}^D(c)$};
        }
        
        \node[prob] (q3d) [above=of ddr3.north -| ddr5.east, anchor=south east] {
            \begin{tabular}{rl}
                \multicolumn{2}{c}{$\bm{q}_3^D(c)$}\\
                \texttt{apple$_1$}: & 0.0426\\
                \texttt{apple$_2$}: & 0.9084\\
                \texttt{apple$_3$}: & 0.0403\\
            \end{tabular}
        };
        \node[shape=coordinate] (int2) [above=2.5mm of ddr3.north] {};
        \draw (ddr3.north) [rounded corners] -- (int2) -- (int2 -| q3d) [arrow] -- (q3d.south);
        \node[loss] (jm) [below=2.5mm of p3.south |- q3p.north, anchor=north] {$J^M$};
        \node[loss] (jd) [above=2.5mm of p3.south |- q3p.south, anchor=south] {$J^D$};
        \draw[arrow] (q3p.east |- jm.west) -- (jm.west);
        \draw[arrow] (q3p.east |- jd.west) -- (jd.west);
        \draw[arrow] (q3d.west |- jm.east) -- (jm.east);
        \node[comment] [left=1mm of q3p.west |- q3p.south, anchor=south east] {
            \begin{tabular}{c}
                $J^M$ encourages prediction\\
                and disambiguation sense\\
                probabilities to match,\\
                $J^D$ encourages only one\\
                sense of target word to\\
                have high probability\\
            \end{tabular}
        };
        \node[comment] [left=1mm of ddr1] {
            \begin{tabular}{c}
                Disambiguation\\
                outputs for $c$\\
            \end{tabular}
        };
        
        \node[transformer] (dlayer2) [below=of ddr3] {Disambiguation Transformer Encoder, $C^D$};
        
        \foreach \step in {1,...,5} {
            \draw[arrow] (dlayer2.north -| ddr\step.south) -> (ddr\step.south);
            \node [vec] (die\step) [below=of dlayer2.south-|ms\step] {$\bm{x}(w_{\step})$};
            \draw[arrow] (die\step.north) -> (dlayer2.south -| die\step.north);
        }
        \node[comment] [left=1mm of die1] {Input embeddings for $c$};
        
        \foreach \i [count=\step from 1] in {i,like,apple,pie,.} {
            \node [token] (dos\step) [below=of die\step] {\i};
            \draw[arrow] (dos\step.north) -> (die\step.south);
        }
        \node[comment] [left=1mm of dos1] {Unmasked sequence $c$};
    \end{tikzpicture}}
    \caption{Architecture diagram for PolyLM when training, illustrated on the sentence ``I like apple pie.'', where the word ``apple'' is chosen as a target and masked (note that ``apple'' is ambiguous when tokens are lower-cased, as it may refer to a fruit or a technology company). At inference time, the bottom components (up to and including $\bm{q}^D(c)$) do not need to be evaluated, and the sequence may not be masked at the input.}
    \label{fig:polylm architecture}
\end{figure*}
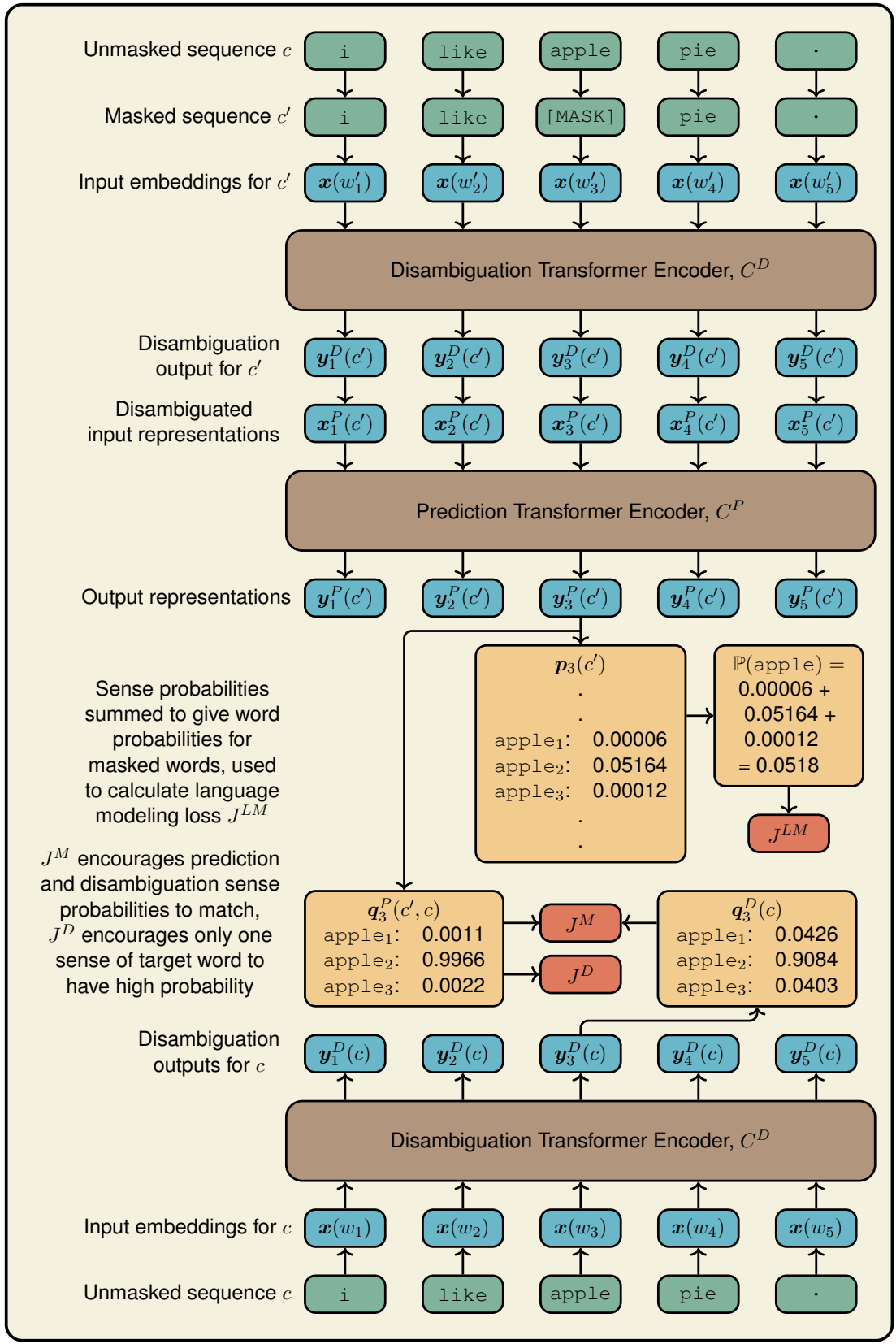

\subsection{Input Layer}
We define a contextualizer to be a function which maps a sequence of input representations $\bm{x}_1, \bm{x}_2, ..., \bm{x}_n \in \mathbb{R}^d$ to a corresponding sequence of output representations $\bm{y}_1, \bm{y}_2, ..., \bm{y}_n \in \mathbb{R}^d$. Recurrent Neural Networks and Transformer architectures are both commonly used as contextualizers for language modeling. Typically the input representations are drawn from an embedding matrix $I \in \mathbb{R}^{|V| \times d}$. It has become common (e.g. BERT) to set $I$ equal to $O$, the embedding matrix used at the language modeling output, as recommended by \citet{press-wolf-2017-using}, and thus have a single embedding matrix $E$.

The issue of input representation poses a problem for our model. Our output embeddings $E \in \mathbb{R}^{|S| \times d}$ correspond to sememes. We cannot straightforwardly tie our input and output embeddings as \citeauthor{press-wolf-2017-using} suggest, because we receive words rather than sememes as input. We solve this problem by setting the input representation of a word to be a convex combination of the representations of its sememes, i.e.
\begin{align} \label{eq:sememe to word representations}
    \bm{x}(w) = \sum_{s \in S_w} \lambda_{ws} \bm{e}_s,
\end{align}
where $\bm{e}_s$ is the row of $E$ corresponding to sememe $s$, and $\bm{\lambda}_w$ is a learnable weight vector with the properties that $\sum_{s \in S_w} \lambda_{ws} = 1$ and $\bm{\lambda}_w \ge \bm{0}$ (in practice, $\bm{\lambda}_w$ is the softmax of an underlying, unconstrained variable vector).

\subsection{Disambiguation Layer}
The disambiguation layer attempts to infer the contextually appropriate sememe embeddings for the input based on the conflated representations from the input layer.

Representations $\bm{x}(w'_1), \bm{x}(w'_2), ..., \bm{x}(w'_n)$ of $c'$, calculated according to Eq. \ref{eq:sememe to word representations}, are fed into a contextualizer instance $C^D$, which outputs representations $\bm{y}_1^D(c'), \bm{y}_2^D(c'), ..., \bm{y}_n^D(c')$. We use these representations to calculate a probability distribution over each sense of the tokens in the input:
\begin{align} \label{eq:disambiguation sense probs}
    \bm{q}_i^D(c') = \text{softmax}\big(E^{(w'_i)} \bm{y}_i^D(c') + \bm{b}^{(w'_i)}\big),
\end{align}
where $E^{(w'_i)}$ is a submatrix of $E$ containing only the rows corresponding to senses of token $w'_i$, and similarly $\bm{b}^{(w'_i)}$ is a subvector of a learnable bias vector $\bm{b} \in \mathbb{R}^{|S|}$. In other terms,
\begin{align}
    q_{is}^D(c') = \frac{e^{\bm{e}_s^\top \bm{y}_i^D(c') + b_s}}{\sum\limits_{s' \in S_{w'_i}} e^{\bm{e}_{s'}^\top \bm{y}_i^D(c') + b_{s'}}},
\end{align}
where $s \in S_{w'_i}$. $q_{is}^D(c')$ corresponds to the probability that the $i$th token in sequence $c'$ has sense $s$.

The disambiguated representation of a token could simply be its highest-probability sememe embedding in the context, but to allow gradients to flow through the disambiguation layer, we take the sum of the sememe embeddings weighted by their probabilities:
\begin{align} \label{eq:disambiguated representations}
    \bm{x}_i^P(c') = \sum_{s \in S_{w'_i}} q_{is}^D(c') \bm{e}_s.
\end{align}

\subsection{Prediction Layer}
The prediction layer maps a sequence of disambiguated input representations onto a corresponding set of output representations, and from each output representation estimates the probability of every sememe in the sense inventory occurring at the corresponding position of the sequence.

Disambiguated representations $\bm{x}_1^P(c'), \bm{x}_2^P(c'), ..., \bm{x}_n^P(c')$ are fed into another contextualizer instance $C^P$, which returns output representations $\bm{y}_1^P(c'), \bm{y}_2^P(c'), ..., \bm{y}_n^P(c')$. These are used to calculate a probability distribution over the entire sense inventory, as prescribed by Eq. \ref{eq:sense probability vector}:
\begin{align} \label{eq:prediction softmax output}
    \bm{p}_i(c') = \text{softmax}(E \bm{y}_i^P(c') + \bm{b}).
\end{align}

We define an additional set of probabilities $\bm{q}^P$ analogous to $\bm{q}^D$ defined in Eq. \ref{eq:disambiguation sense probs}:
\begin{align}
    \bm{q}_i^P(c', c) = \text{softmax}\big(E^{(w_i)} \bm{y}_i^P(c') + \bm{b}^{(w_i)}\big).
\end{align}

$\bm{q}_i^P$ takes both $c'$ and the unmasked sequence $c$ as arguments because we are interested in the sense probabilities of the words $w_i$ that actually occurred. $\bm{q}_i^P$ will be used later for defining the loss function and is useful for downstream tasks.

\subsection{Loss Function}
We seek to minimize a loss function $J$ with three components, each of which is explained below:
\begin{equation}
\begin{split}
    J(c, c', T) =\ &J^{LM}(c, c', T)\ +\\
                         &J^{D}(c, c', T)\ +\\
                         &J^{M}(c, c', T)
\end{split}
\end{equation}

\subsubsection{Language Modeling Loss}
The language modeling loss $J^{LM}$ is defined as the mean negative log likelihood of the target tokens occurring:
\begin{align}
    &J^{LM}(c, c', T) \nonumber\\
    &\quad= \frac{-1}{|T|} \sum_{i \in T} \log \hat{\pr}(w_i\ |\ c')\\
    &\quad= \frac{-1}{|T|} \sum_{i \in T} \log \sum_{s \in S_{w_i}} \hat{\pr}(\text{sememe } i \text{ is } s\ |\ c')\\
    &\quad= \frac{-1}{|T|} \sum_{i \in T} \log \sum_{s \in S_{w_i}} p_{is}(c^\prime),
\end{align}
where $\bm{p}_i$ is as defined in Eq. \ref{eq:prediction softmax output}.

\subsubsection{Distinctness Loss}
Recall that we assume in advance a number of senses for each word. In practice we guess a relatively high number to avoid missing senses. When we overestimate the number of senses, we find that two different sense embeddings for a word converge to essentially the same meaning. The aim of the distinctness loss is to ensure that each sense has a distinct meaning, and to ``kill off'' superfluous senses by causing them to have very low probability in all contexts.

The second key observation of PolyLM is that if the sememes corresponding to a word $w$ are distinct, then in contexts where $w$ occurs, we would expect one of these sememes to have a high estimated probability of occurring, and the rest to have a low probability. The distinctness loss, given by,
\begin{align}
    &J^D(c, c', T) = \frac{-1}{r|T|} \sum_{i \in T} \log \sum_{s \in S_{w_i}} \big(q_{is}^P(c', c)\big)^r,
\end{align}
with hyperparameter $r > 1$, encourages this separation to occur. A full justification is given in Appendix \ref{sec:distinctness loss derivation}.

\subsubsection{Match Loss}
Without extra supervision, the disambiguation layer tends to very quickly allocate almost all of the probability mass for a word to a single one of its senses. This appears to be due to a ``rich get richer'' effect in Eq. \ref{eq:disambiguated representations}, where the sense embedding with the highest weight has larger gradients associated with it.

A more reliable source of sense probabilities is the output of the prediction layer, as this is more closely associated with the ground truth. Therefore we encourage the disambiguation sense probabilities $\bm{q}^D$ to be similar to the prediction sense probabilities $\bm{q}^P$ by adding a sense probability ``match loss,'' which is proportional to the cosine similarity between $\bm{q}^D$ and $\bm{q}^P$.

Because $\bm{q}_i^D(c')$ is meaningless when token $i$ is replaced with [MASK], when calculating the match loss we evaluate the disambiguation layer on the unmasked sequence (shown with bottom-up arrows in Figure \ref{fig:polylm architecture}), obtaining $\bm{q}_i^D(c)$. The match loss is defined as
\begin{align}
    J^M(c, c', T) = \frac{-\lambda^M}{|T|} \sum_{i \in T} \frac{\bm{q}_i^D \cdot \bm{q}_i^P}{\|\bm{q}_i^D\| \|\bm{q}_i^P\|},
\end{align}
where $\bm{q}_i^D$ and $\bm{q}_i^P$ are shorthand for $\bm{q}_i^D(c)$ and $\bm{q}_i^P(c', c)$ respectively, and $\lambda^M$ is a hyperparameter.

As we wish the disambiguation layer to learn from the prediction layer rather than the other way around, we do not allow gradients from the match loss to propagate through $\bm{q}_i^P$.

\subsection{Details and Parameters}
\subsubsection{Preprocessing}
To avoid the issue of how to represent a word's sense when it is broken into sub-word level tokens, our vocabulary consists of whole-word tokens. However the WSI tasks on which we evaluate our model operate on the lemma level, so we lemmatize our training corpus as described in Appendix \ref{sec:lemmatization}. The vocabulary consists of the $\roughly$86K tokens appearing more than 500 times in our training corpus, which like BERT's consists of English Wikipedia + BookCorpus \cite{zhu-etal-2015-aligning}. All tokens are lower-cased.

\subsubsection{Contextualizers}
One of the advantages of PolyLM is that it can be used with any type of contextualizer - note however that we must train our contextualizers together with the rest of the model rather than using pretrained contextualizer instances, because their word embedding matrix would not match our sense embedding matrix. In this paper we present results where the disambiguation and prediction contextualizers $C^D$ and $C^P$ use BERT's implementation of the Transformer encoder architecture.

\subsubsection{Parameters}
To keep the total number of embeddings reasonable, we allow only the $\roughly$10,000 tokens which occur more than 20,000 times in the training corpus, or appear as focuses in the evaluation datasets, to have multiple senses. Specifically, we assign these tokens a fixed number of $k = 8$ embeddings, and other tokens a single embedding. Since according to Zipf's law \cite{zipf-1950-human}, it is the most frequent words which tend to have the most senses, we expect not to miss too many senses by assuming that infrequent words are monosemous. We leave the investigation of more sophisticated methods for pre-allocating or dynamically updating the number of senses for each token for future work.

We train two PolyLM models of different sizes, PolyLM$_\text{SMALL}$ and PolyLM$_\text{BASE}$. Due to the prohibitive computational cost of training a model of BERT$_\text{LARGE}$'s size, we use significantly smaller dimensions, as shown in Table \ref{tab:params}.

Models were trained over 6,000,000 batches consisting of 32 sequences of length 128 using the Adam optimizer \cite{kingma-ba-2014-adam}. The learning rate was increased linearly from 0 to 3e-5 over the first 10,000 batches, and then reduced linearly back to zero over the remaining batches. The hyperparameters $\lambda^M$ and $r$ specific to PolyLM's loss function were first increased linearly and then left constant, $\lambda^M$ from 0 to 0.1 over the first 1,000,000 batches, and $r$ from 1.0 to 1.5 over the first 2,000,000 batches.

It is important for $r$ to be gradually increased in this manner because if $r$ is large initially, then the effect of the distinctness loss reduces the diversity of the senses learned. On the other hand, increasing $r$ too slowly seems to be detrimental to the senses' distinctness.

\begin{table*}[!htb]
    \centering
    \resizebox{\textwidth}{!}{\begin{tabular}{lcccccccc}
        \hline
        \textbf{Model} & $d$ & \textbf{Filter size} & \textbf{No. attn. heads} & \textbf{No. layers} & \textbf{Seq. len.} & \textbf{Vocab size} & \textbf{No. embeddings} & \textbf{Total params}\\
        \hline
        PolyLM$_\text{SMALL}$ & 128 & 512 & 8 & 4 ($C^D$), 8 ($C^P$) & 128 & 86K & 157K & 24M\\
        PolyLM$_\text{BASE}$ & 256 & 1024 & 8 & 4 ($C^D$), 12 ($C^P$) & 128 & 86K & 157K & 54M\\
        BERT$_\text{LARGE}$ & 1024 & 4096 & 16 & 24 & 512 & 30K & 30K & 340M\\
        \hline
    \end{tabular}}
    \caption{Parameters of PolyLM and BERT$_\text{LARGE}$.}
    \label{tab:params}
\end{table*}

\begin{table*}[!htb]
    \centering
    \resizebox{\textwidth}{!}{\begin{tabular}{llcccccc}
        \hline
        \multirow{2}{*}{\textbf{System}} & \multirow{2}{*}{\textbf{Version}} & \multicolumn{3}{c}{\textbf{SemEval-2010}} & \multicolumn{3}{c}{\textbf{SemEval-2013}} \\
        && F-S & V-M & \textbf{AVG} & FBC & FNMI & \textbf{AVG}\\
        \hline
        \citet{amrami-goldberg-2019-towards} & BERT$_\text{LARGE}$ & \textbf{71.3} & 40.4 & \textbf{53.6} & 64.0 & 21.4 & 37.0\\
        AutoSense \cite{amplayo-etal-2019-autosense} & & 62.9 & 10.1 & 25.2 & 61.7 & 8.0 & 22.2\\
        \hline
        \multirow{2}{*}{PolyLM$^{\dagger}$} & BASE & 65.8 & \textbf{40.5} & 51.6 & \textbf{64.8} & \textbf{23.0} & \textbf{38.6}\\
        & SMALL & 65.6 & 35.7 & 48.4 & 64.5 & 18.5 & 34.5\\
        \hline
        \citet{qiu-etal-2016-context}$^\dagger$ & & - & - & - & 56.9 & 6.7 & 19.5\\
        SE-WSI-fix-cmp \cite{song-etal-2016-sense}$^\dagger$ & & 54.3 & 16.3 & 29.8 & - & - & -\\
        AdaGram \cite{bartunov-etal-2016-breaking}$^\dagger$ & & 43.9 & 20.0 & 29.6 & 13.2 & 8.9 & 10.8\\
        \citet{arora-etal-2018-linear}$^\dagger$ & $k = 5$ & 46.4 & 11.5 & 23.1 & - & - & -\\
        \hline
    \end{tabular}}
    \caption{Comparison of sense embedding models and WSI-specific techniques on the SemEval 2010 and 2013 WSI tasks. SE-WSI-fix-cmp is based on \citet{neelakantan-etal-2014-efficient}'s MSSG model. $^\dagger$ - models which obtain explicit sense embeddings.}
    \label{tab:wsi}
\end{table*}

\begin{table*}[!htb]
    \centering
    \begin{tabular}{lcccccc}
        \hline
        \multirow{2}{*}{\textbf{Description}} & \multicolumn{3}{c}{\textbf{SemEval-2010}} & \multicolumn{3}{c}{\textbf{SemEval-2013}} \\
        & F-S & V-M & \textbf{AVG} & FBC & FNMI & \textbf{AVG}\\
        \hline
        PolyLM$_\text{SMALL}$ & \textbf{65.6} & \textbf{35.7} & \textbf{48.4} & \textbf{64.5} & \textbf{18.5} & \textbf{34.5}\\
        No distinctness loss & 53.5 & 33.4 & 42.3 & 57.4 & 16.3 & 30.5\\
        No disambiguation layer & 64.9 & 25.5 & 40.6 & \textbf{64.5} & 17.5 & 33.6\\
        Disambiguation layer only & 63.6 & 29.3 & 43.2 & 62.7 & 15.7 & 31.4\\
        \hline
    \end{tabular}
    \caption{PolyLM ablation study.}
    \label{tab:wsi_ablation}
\end{table*}

\section{Experiments} \label{sec:experiments}
Word sense induction (WSI) is the task of inferring the senses of a word in an unsupervised manner. This is precisely the aim of our method, and so is an ideal test task. We evaluate PolyLM on two WSI datasets, SemEval-2010 Task 14 \cite{manandhar-etal-2010-semeval} and SemEval-2013 Task 13 \cite{jurgens-klapaftis-2013-semeval}. Both datasets consist of passages containing one of a set of polysemous focus words. The occurrences of the focus words in the test set have been sense-labeled by human annotators according to a reference sense inventory.

In the SemEval-2010 dataset, each instance is labeled with a single sense, whereas in the SemEval-2013 dataset an instance may be labeled with several relevant senses, each with a corresponding weight denoting its degree of applicability in the context.

Performance on SemEval-2010 is measured using paired F-Score (F-S) and V-Measure (V-M), and on SemEval-2013 using Fuzzy B-Cubed (FBC) and Fuzzy Normalized Mutual Information (FNMI). Overall performance on each task (\textbf{AVG}) is typically defined as the geometric mean of its two sub-metrics.

Currently, the best performing system on both datasets is that of \citet{amrami-goldberg-2019-towards}. Their system uses the idea of \textit{substitute vectors}, first devised by \citet{baskaya-etal-2013-ai}. For each instance, a set of most likely words that could have occurred instead of the focus word is obtained from the output of a language model. These sets are then clustered, and each cluster is taken to correspond to a different sense of the focus word. \citeauthor{amrami-goldberg-2019-towards} use BERT$_\text{LARGE}$ as their language model.

PolyLM can be used for WSI without any further training. For the SemEval-2010 dataset, each instance $c$ is labeled with the sense of the focus word $w_i$ which has the highest predicted probability, i.e. $\text{argmax}_{s \in S_{w_i}} \bm{q}^P_{is}(c', c)$, where $c'$ is formed from $c$ by replacing $w_i$ with [MASK]. For SemEval-2013, we consider a sense applicable if it has a predicted probability $q^P_{is}(c', c) > p^{\text{thresh}}$, and the weight assigned to each applicable sense is its probability $q^P_{is}(c', c)$. We arbitrarily set $p^{\text{thresh}}$ to 0.2.

Results are shown in Table \ref{tab:wsi}. Both PolyLM models comprehensively outperform previous sense embedding methods. PolyLM$_\text{BASE}$ and \citeauthor{amrami-goldberg-2019-towards}'s system slightly outperform each other on one dataset each, suggesting similar overall proficiency at WSI. However it is worth noting that the BERT$_\text{LARGE}$ language model used by \citeauthor{amrami-goldberg-2019-towards} has more than six times as many parameters as PolyLM$_\text{BASE}$ and is much more computationally expensive to train and run.

PolyLM scales well for the sizes tested, with PolyLM$_\text{BASE}$ outperforming PolyLM$_\text{SMALL}$ by 3.2 and 4.1 points in \textbf{AVG} score on the two datasets with a 2.25x increase in the number of parameters. Even if further increases in model dimensions yielded much smaller improvements in performance, it seems likely that a PolyLM model of BERT$_\text{LARGE}$'s 340 million parameter size would achieve results significantly better than those of \citet{amrami-goldberg-2019-towards}.

\subsection{Ablation Study}

We test three alternative configurations against PolyLM$_\text{SMALL}$: one where the distinctness loss term is removed from the objective (``no distinctness loss''), one where the disambiguation layer is removed (``no disambiguation layer''), and one where the disambiguation sense probabilities $\bm{q}^D$ are used in place of $\bm{q}^P$ when performing WSI (``disambiguation layer only''). Note that the first two configurations require new models to be trained, whereas the last simply uses PolyLM$_\text{SMALL}$ in a different way. Results are shown in Table \ref{tab:wsi_ablation}.

The use of the distinctness loss has a big impact on model performance, while the disambiguation layer is somewhat less important but still useful. The model still performs surprisingly well when the disambiguation rather than the prediction sense probabilities are used; these are the output of only four Transformer layers and hence are much cheaper to compute. This suggests that it might be practical to add the disambiguation layer at the input of various neural NLP models to improve their understanding of polysemy.

\section{Conclusions} \label{sec:discussion}
PolyLM is a novel model of polysemy based on two assumptions about word senses: firstly, that the probability of a word occurring in a context is equal to the sum of its individual senses occurring, as expressed by the language modeling loss; and secondly, that generally only one sense of a word ought to have a high probability of occurring in a given context, as expressed by the distinctness loss. PolyLM does indeed learn word senses which correspond well to human notions, as demonstrated by its performance on word sense induction, which matches that of the previous state-of-the-art system despite having 6 times fewer parameters. It can be easily applied to many word-sense related tasks, as it generates a probability distribution over the senses of each word in the input text. It is not specific to any one contextualizer and so can be improved as contextualizers improve.

\section*{Acknowledgements}
Felipe Bravo-Marquez was funded by ANID FONDECYT grant 11200290, U-Inicia VID Project UI-004/20 and ANID - Millennium Science Initiative Program - Code ICN17\_002.

\bibliography{anthology,references}
\bibliographystyle{acl_natbib}

\appendix

\section{Justification of the Distinctness Loss} \label{sec:distinctness loss derivation}
Consider the derivative of the language modeling loss for one particular target position $i \in T$ with respect to the pre-softmax scores $\bm{e}_k^\top \bm{y}_i^P + b_k$ of the target word $w_i$'s sense embeddings $k \in S_{w_i}$. For brevity, we define $y_k = \bm{e}_k^\top \bm{y}_i^P + b_k$.
\begin{align*}
    &-\pd{}{y_k} J^{LM}(c, c', \{i\})\\
    &\ = \pd{}{y_k} \log \sum_{s \in S_{w_i}} \big[\text{softmax}(E \bm{y}_i^P + \bm{b})\big]_s\\
    &\ = \pd{}{y_k} \log \sum_{s \in S_{w_i}} \frac{e^{y_s}}{\sum_{s' \in S} e^{y_{s'}}}\\
    &\ = \pd{}{y_k} \log \frac{\sum_{s \in S_{w_i}} e^{y_s}}{\sum_{s \in S_{w_i}} e^{y_s} + \sum_{s \in S \setminus S_{w_i}} e^{y_s}}\\
    &\ = \pd{}{y_k} \log \frac{\sum_{s \in S_{w_i}} e^{y_s}}{\sum_{s \in S_{w_i}} e^{y_s} + C},\\
    &\text{where } C = \sum_{s \in S \setminus S_{w_i}} e^{y_s},\\
    &\ = \pd{}{y_k}\bigg(\log \sum_{s \in S_{w_i}} e^{y_s} - \log \Big(\sum_{s \in S_{w_i}} e^{y_s} + C\Big)\bigg)\\
    &\ = \frac{\pd{}{y_k} \sum_{s \in S_{w_i}} e^{y_s}}{\sum_{s \in S_{w_i}} e^{y_s}} - \frac{\pd{}{y_k} (\sum_{s \in S_{w_i}} e^{y_s} + C)}{\sum_{s \in S_{w_i}} e^{y_s} + C}\\
    &\ = \frac{e^{y_k}}{\sum_{s \in S_{w_i}} e^{y_s}}  - \frac{e^{y_k}}{\sum_{s \in S_{w_i}} e^{y_s} + C}\\
    &\ = \frac{e^{y_k}}{\sum_{s \in S_{w_i}} e^{y_s}}  - \frac{e^{y_k}}{\sum_{s \in S} e^{y_s}}\\
    &\ = q_{ik}^P(c', c) - p_{ik}(c).
\end{align*}
Since $q_{ik}^P > p_{ik}$, $\pd{}{y_k} J^{LM}(c, c', \{i\})$ will always be negative, meaning that every sense embedding for the target word will always move towards the contextualized representation $\bm{y}_i^P$. This is undesirable, because it means that even senses which are irrelevant in a context will receive a positive update.

Now consider the derivatives of the distinctness loss:
\begin{align*}
    &-\pd{}{x_y} J^D(c, c', \{i\})\\
    &\ = \pd{}{y_k} \frac{1}{r} \log \sum_{s \in S_{w_i}} \big(q_{is}^P(c', c)\big)^r\\
    &\ = \frac{1}{r} \pd{}{y_k} \log \sum_{s \in S_{w_i}} \bigg(\frac{e^{y_s}}{\sum_{s' \in S_{w_i}} e^{y_{s'}}}\bigg)^r\\
    &\ = \frac{1}{r} \pd{}{y_k} \log \sum_{s \in S_{w_i}} \frac{e^{ry_s}}{(\sum_{s' \in S_{w_i}} e^{y_{s'}})^r}\\
    &\ = \frac{1}{r} \pd{}{y_k} \log \frac{\sum_{s \in S_{w_i}} e^{ry_s}}{(\sum_{s \in S_{w_i}} e^{y_s})^r}\\
    &\ = \frac{1}{r} \pd{}{y_k} \bigg(\log \sum_{s \in S_{w_i}} e^{ry_s} - \log \Big(\sum_{s \in S_{w_i}} e^{y_s}\Big)^r\bigg)\\
    &\ = \frac{1}{r} \bigg(\frac{\pd{}{y_k} \sum_{s \in S_{w_i}} e^{ry_s}}{\sum_{s \in S_{w_i}} e^{ry_s}} - \frac{r \pd{}{y_k} \sum_{s \in S_{w_i}} e^{y_s}}{\sum_{s \in S_{w_i}} e^{y_s}}\bigg)\\
    &\ = \frac{1}{r} \bigg(\frac{re^{ry_k}}{\sum_{s \in S_{w_i}} e^{ry_s}} - \frac{r e^{y_k}}{\sum_{s \in S_{w_i}} e^{y_s}}\bigg)\\
    &\ = \frac{e^{ry_k}}{\sum_{s \in S_{w_i}} e^{ry_s}} - q_{ik}^P(c', c).
\end{align*}
When $r > 1$, $\frac{e^{ry_k}}{\sum_{s \in S_{w_i}} e^{ry_s}}$ is a ``sharpened'' version of $q_{ik}^P(c', c)$: it is larger than $q_{ik}^P$ when $q_{ik}^P$ is large, and smaller when $q_{ik}^P$ is small.

Now we have
\begin{align*}
    &-\pd{}{y_k} \Big(J^{LM}(c, c', \{i\}) + J^D(c, c', \{i\})\Big)\\
    &\ = -\pd{}{y_k} J^{LM}(c, c', \{i\}) - \pd{}{y_k} J^D(c, c', \{i\})\\
    &\ = q_{ik}^P(c', c) - p_{ik}(c) + \frac{e^{ry_k}}{\sum_{s \in S_{w_i}} e^{ry_s}} - q_{ik}^P(c', c)\\
    &\ = \frac{e^{ry_k}}{\sum_{s \in S_{w_i}} e^{ry_s}} - p_{ik}(c)\\
    &\ = q_{ik}^{\text{sharp}}(c', c) - p_{ik}(c).
\end{align*}
Thus the addition of the distinctness loss results in even stronger reinforcement for senses which are highly applicable in the context, and even weaker (possibly negative) reinforcement for senses which are inapplicable. This encourages only one sense of a word to have high probability in a given context, as desired.

\section{Lemmatization} \label{sec:lemmatization}
The training corpus and all text used for evaluation are lemmatized as follows: first, we perform part-of-speech (POS) tagging using Stanford CoreNLP's POS tagger \cite{manning-etal-2014-stanford}. Any token with a tag associated with inflectional morphology in English (NNS, JJR, JJS, RBR, RBS, VBD, VBG, VBP, VBZ or VNB) is split into two separate tokens, its lemmatized form and a special token. There is a unique special token for each of the above tags except the pairs JJR and RBR (comparative adjectives and adverbs) and JJS and RBS (superlative adjective and adverbs), which share [COMP] and [SUP] tokens respectively. 

\end{document}